\pdfoutput=1

\documentclass[11pt]{article}

\usepackage[preprint]{acl}

\usepackage{times}
\usepackage{latexsym}

\usepackage[T1]{fontenc}

\usepackage[utf8]{inputenc}

\usepackage{microtype}

\usepackage{inconsolata}

\usepackage{graphicx}
\usepackage{array}
\usepackage{hhline}
\usepackage{multicol}
\usepackage{booktabs}
\usepackage{xcolor}

\title{GRAD-SUM: Leveraging Gradient Summarization for Optimal Prompt Engineering}

\author{%
  Derek Austin\thanks{Primary author.} \\
  Galileo Technologies \\
  \And
  Elliott Chartock \\
  Galileo Technologies \\
}
\begin{document}

\maketitle
\begin{abstract}
Prompt engineering for large language models (LLMs) is often a manual time-intensive process that involves generating, evaluating, and refining prompts iteratively to ensure high-quality outputs.
While there has been work on automating prompt engineering, the solutions generally are either tuned to specific tasks with given answers or are quite costly.
We introduce GRAD-SUM, a scalable and flexible method for automatic prompt engineering that builds on gradient-based optimization techniques. 
Our approach incorporates user-defined task descriptions and evaluation criteria, and features a novel gradient summarization module to generalize feedback effectively.
Our results demonstrate that GRAD-SUM consistently outperforms existing methods across various benchmarks, highlighting its versatility and effectiveness in automatic prompt optimization.
\end{abstract}

\begin{table*}[t]
    \centering
    \renewcommand{\arraystretch}{1.5}
    \setlength{\tabcolsep}{10pt}
    \begin{tabular}{|l|p{0.75\textwidth}|}
        \hline
        \textbf{Prompt State} & \textbf{Prompt} \\
        \hline
        Initial Prompt & Answer the question: \{question\} \\
        \hline
        Optimized Prompt &You are tasked with solving a grade school math question. Follow these detailed steps to ensure a clear and accurate solution: \newline \newline 1. Identify the Mathematical Operation: Begin by determining the type of math operation required (e.g., addition, subtraction, multiplication, division). Clearly state this in your response and explain why this operation is needed for the given question. \newline \newline 2. Outline the Thought Process: Provide a logical and structured approach to solving the problem. Explain your reasoning in detail, as if you are teaching the concept to a student. Ensure each step is connected and follows logically from the previous one, making it easy to follow. \newline \newline 3. Step-by-Step Calculations: Break down the problem into smaller, manageable steps. Show each calculation in detail to demonstrate how you arrive at the solution. Include all intermediate steps and results to offer a comprehensive understanding.\newline \newline 4. Verify Accuracy: After arriving at the solution, double-check your calculations and the final numerical answer to ensure its accuracy. Explain how you verified the correctness of your answer, such as by using inverse operations or checking with a different method.\newline \newline 5. Final Answer: Present the final answer in a clear and precise manner, ensuring it is easy to understand. State explicitly that the answer has been verified for accuracy.\newline \newline Here is the question: \{question\}\\
        \hline
    \end{tabular}
    \caption{Example GRAD-SUM optimization run using the GSM8K dataset.}
    \label{tab:prompt_comparison}
\end{table*}

\section{Introduction}
The introduction of machine learning and artificial intelligence has automated many tasks, but creating and refining prompts for LLM applications remains largely a manual process.
Practitioners often struggle to find the right prompt to elicit accurate and high-quality outputs. 
Common practice is to judge output quality by implementing fully automated evaluation feedback loops, where LLMs assess outputs based on user-provided evaluation criteria, these are often called LLM-as-a-judge methods.
Seeing as evaluation is an automatic process in LLM-as-a-judge scenarios, the main bottleneck in automating prompt engineering is the actual refinement of the prompt.
Although there have been attempts to automate this process \citep{llmAsOptimizers} \citep{auto_prompt_optim_with_gradient_descent} \citep{human_level_prompt_engineers}, we found that these methods either could not be adapted seamlessly to all tasks or were too costly to implement at scale.

Automatic prompt optimization methods broadly fall into two categories: Monte Carlo search methods \cite{llmAsOptimizers} and feedback-based (or gradient-based) methods \citep{auto_prompt_optim_with_gradient_descent}\cite{promptbreeder}\citep{grips}.
Monte Carlo search methods start with a base prompt and iterate over hundreds if not thousands of possible prompts in order to find the best performer, an inefficient process.
These methods use minimal feedback on why their responses might be incorrect and instead search somewhat blindly through prompt space to identify the best possible prompt, making small changes along the way.
Feedback-based methods, on the other hand, use ratings and explanations of how their answer could be improved from previous iterations to generate new prompts. 
They aim to emulate a traditional optimization process by iteratively refining prompts based on detailed feedback, closely mimicking the approach of a human prompt engineer.
As shown in \citep{llmAsOptimizers}, Monte Carlo methods take many iterations to converge, and thus are likely to be too costly for most practitioners.
Therefore, in approaching the problem of automatic prompt engineering we choose to focus on feedback-based methods.

\begin{figure*}[ht]
    \centering
    \hspace*{-.001cm}\includegraphics[width=1\textwidth]{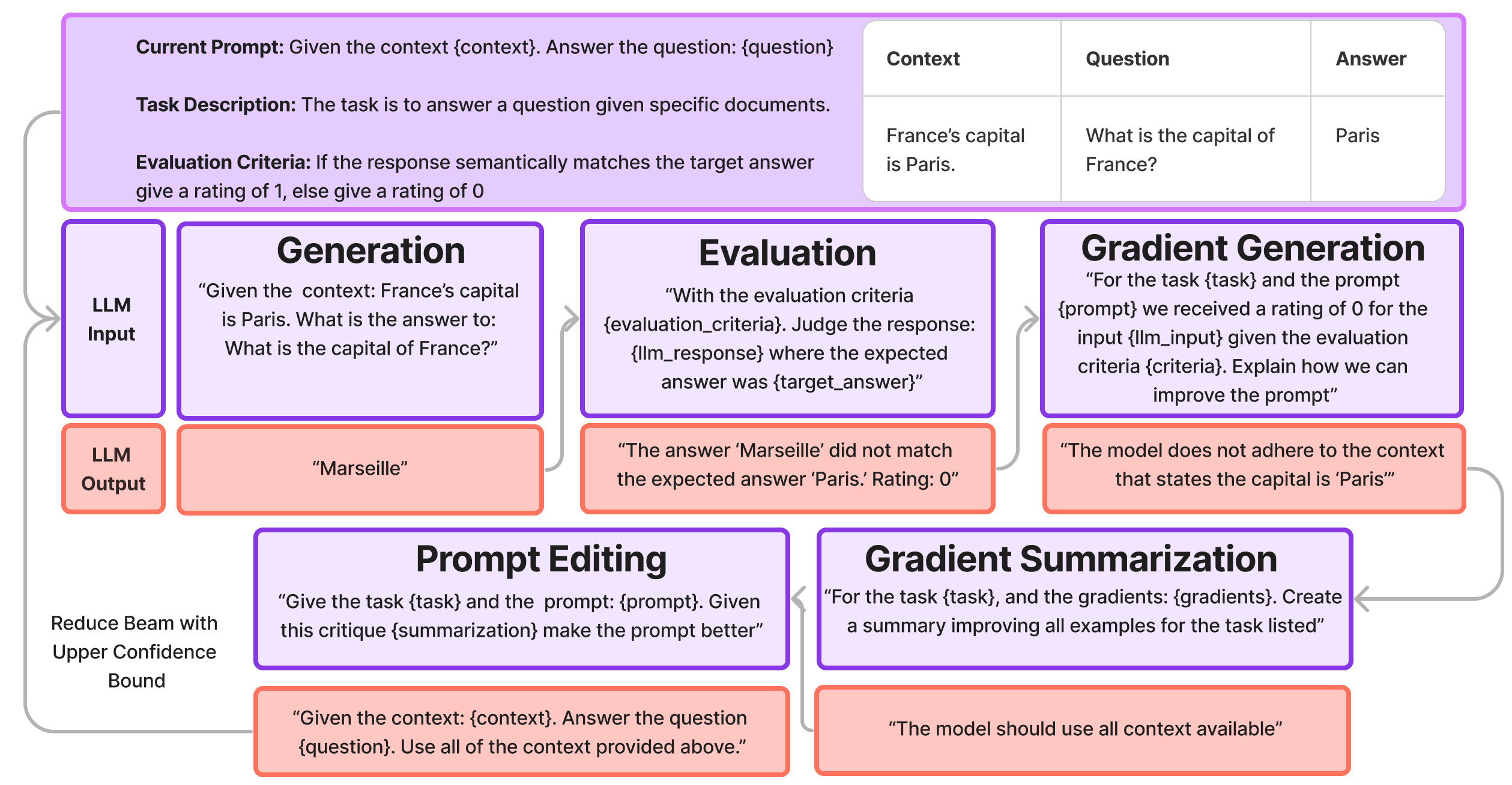}
    \caption{An illustration of one GRAD-SUM training loop. Modules are sequential starting with generation. The prompt chosen in our prompt editor model is then fed back to the generation module and the training loop restarts.}
    \label{fig:diagram}
\end{figure*}

One of the seminal works that inspired our feedback-based approach was presented in "Automatic Prompt Optimization with `Gradient Descent'
and Beam Search" \citep{auto_prompt_optim_with_gradient_descent}. 
`Gradients' in this context, refers to natural language descriptions of errors in LLM responses and suggestions for improving the system's response in the future.
The method presented in the aforementioned paper uses each of these gradients individually to generate a new prompt.
Their approach addressed classification tasks such as hate speech detection, lie detection, jailbreak detection, and sarcasm detection, using a restrictive matching evaluator where predictions must exactly match the expected answers.
As most LLM use cases do not have a strict expected output we improve upon their work by introducing LLMs-as-a-judge evaluations - which enable accurate judgment of any response for all possible tasks.

Our approach, GRAD-SUM, introduces a task description input that allows models to incorporate high-level, domain-specific information provided by the user. 
Additionally, we introduce natural language evaluation criteria also provided by the user, enabling our method to use LLM-as-a-judge metrics, extending \citep{auto_prompt_optim_with_gradient_descent} method to all possible use cases.
We also improve upon \citep{auto_prompt_optim_with_gradient_descent} method, by introducing a new gradient summarization module after discovering that editing the prompt based on feedback from a single output, as done in \citep{auto_prompt_optim_with_gradient_descent} often resulted in highly specific candidates prompts that failed to generalize to the broader dataset population. 
To address this, our gradient summarization module takes all computed gradients and produces a generalized summary of the feedback, akin to averaging gradients over many samples, a common practice in machine learning. 
Overall, our approach combines a novel gradient summarization module, user-provided task descriptions and evaluation criteria to create a more flexible, scalable and performant solution for automated prompt engineering than any existing today.

\section{Related Work}
Customizing an LLM to a user's specific task can fit into two broad categories: parametric and non-parametric approaches.
Parametric approach's usually consist of parameter efficient fine-tuning \citep{lora} \citep{prefix_tuning} where only a portion of parameters are updated based on backpropagating gradients.
Prefix tuning introduces new learnable tokens that one can think of as a prompt in continuous space, that are learned for specific tasks \citep{prefix_tuning}.
However, as these tokens defined in continuous space they are not easily interpretable.
Other approaches like \citep{autoprompt} backpropagate gradients and choose the best discrete prompts. 
However, even this technique suffers from interpretability as most prompts are not grammatically correct or coherent sentences.

Most practitioners currently rely on black box API based LLM's for language generation where there is no potential for backpropagation of any kind \citep{gpt3}.
Therefore, prompt optimization becomes a discrete optimization problem over an extraordinarily large search space.
Some methods attempt to tackle this problem through the use of complementary models.
\citep{rlprompt} attempt to use a policy network that receives rewards from an evaluator function in order to update a discrete prompt.
\citep{instructzero} uses an open source model to convert a soft-prompt defined in embedding space into a discrete prompt that was then fed to the LLM API.
Others use evolutionary algorithms in order to modify, delete and add a list of candidate prompts that are evaluated each iteration \citep{promptbreeder} \citep{grips}.
Simple Monte Carlo sampling around candidate prompts \citep{llmAsOptimizers} can complement these methods or be implemented entirely on their own.

Most of these methods attempt to use some sort of signal from an evaluator LLM in order to asses the quality of the output.
However, it has been shown that LLMs are not great at self-correction without use of auxiliary knowledge or tools \citep{gpt4_does_not_know_its_wrong} \citep{are_llms_good_optimizers} \citep{LLM_cannot_correct_self_reasoning} \citep{critic}.
Therefore the introduction of specific evaluation criteria, as in our method, the use of tools, as in \citep{critic}, or modifying LLM critiques to better find the root of the errors \citep{are_llms_good_optimizers} becomes of paramount importance to finding an optimized prompt.

\section{Method}
Our method consists of 5 distinct modules as seen in Figure \ref{fig:diagram}: generation, evaluation, gradient generation, gradient summarization, and prompt editing.
Throughout our optimization process, we utilize beam search with a beam size of three candidate prompts, feeding each potential prompt to the modules introduced below. Figure \ref{fig:diagram} showcases one iteration for one potential prompt.

\subsection{Generation}
The generation module necessitates a dataset, a prompt, and an LLM with which to generate outputs.
In usual workflows the prompt will consist of formatting `slots' where the dataset will provide text that will be used to fill in the slots as shown in Figure \ref{fig:diagram}.
For instance, a prompt for a retrieval augmented generation (RAG) workflow could be:
\begin{quote}
``Given the following context: \{context\}. Answer the question \{question\}."
\end{quote}
The dataset should then consist of the columns `context', `question' and optionally target outputs which can be used in the evaluate module.
We then feed these formatted prompts to the generation function and receive LLM generated responses.

\subsection{Evaluation}
Our evaluation module then takes in the generations from the generation function, user-defined evaluation criteria, and optionally an expected answer.
The module in turn returns a rating as to how well the response performs as judged by the evaluation criteria as well as an explanation for the rating.
We ask for the explanation before the evaluator module returns a rating as chain of thought has been shown to dramatically improve LLM performance, especially on reasoning tasks \citep{chain_of_thought}.
For use cases with an expected answer, evaluation criteria can be as simple as:
\begin{quote}
``Does the LLM generated response semantically match the expected answer? If a reasonable human would judge it to, give the response a 1, otherwise a 0."
\end{quote}
We note that we only ask for a binary indicator as empirically we found LLM's are far better and more consistent at binary indicators than a sliding numeric scale.

\subsection{Gradients}
The gradient module receives the current prompt, the evaluation criteria, a description of the task at hand, and a maximum of 5 generation responses that received a rating of 0 from the evaluation module. 
Note that any generation receiving a rating of 1 in the evaluate step will be excluded from this process as it cannot be improved, thus it is possible the gradient module will receive less than 5 inputs.
An LLM is then asked to evaluate the input and output for the specific prompt and provide actionable methods for improving the prompts to address deficiencies noted in the rating explanation. 
The gradient module leverages all available information—the task description, input to the generation function, the LLM response from the generate function, evaluation criteria, and explanations for poor ratings—to identify areas for improvement.
Below is a sample of what a gradient will look like:
\begin{quote}
    ``The model should draw upon the entire context and state its reasoning explicitly."
\end{quote}
This step aims to emulate human evaluation of generations, determining how to adjust the prompt to achieve the desired outcomes.

\begin{table*}[h]
    \centering
    \renewcommand{\arraystretch}{1.2}
    \resizebox{\textwidth}{!}{
    \begin{tabular}{|l|l|l|c|c|}
        \hline
        Dataset & Generating Model & Optimization Method & Initial Validation Rating & Final Validation Rating \\
        \hhline{=====}
        GSM8K & GPT 3.5 & DSPY & 0.635 & 0.755 \\
        \hline
        GSM8K & GPT 3.5 & GRAD-SUM & 0.635 & \textbf{0.82} \\
        \hhline{=====}
        Orca Math & GPT 3.5 & DSPY & 0.395 & 0.455 \\
        \hline
        Orca Math & GPT 3.5 & GRAD-SUM & 0.395 & \textbf{0.575} \\
        \hhline{=====}
        Neural Bridge RAG & GPT 3.5 & DSPY & 0.605 & 0.885 \\
        \hline
        Neural Bridge RAG & GPT 3.5 & GRAD-SUM & 0.605 & \textbf{0.915} \\
        \hhline{=====}
        HellaSwag & GPT 3.5 & DSPY & 0.575 & 0.48 \\
        \hline
        HellaSwag & GPT 3.5 & GRAD-SUM & 0.575 & \textbf{0.795} \\
        \hhline{=====}
        HotPot QA & GPT 3.5 & DSPY & 0.575 & 0.626 \\
        \hline
        HotPot QA & GPT 3.5 & GRAD-SUM & 0.575 & \textbf{0.725} \\
        \hhline{=====}
        MMLU & GPT 3.5 & DSPY & 0.45 & 0.56 \\
        \hline
        MMLU & GPT 3.5 & GRAD-SUM & 0.45 & \textbf{0.625} \\
        \hhline{=====}
        MT \& Vicuna Bench & GPT 3.5 & DSPY & 0.831 & 0.823 \\
        \hline
        MT \& Vicuna Bench & GPT 3.5 & GRAD-SUM & 0.831 & \textbf{0.95} \\
        \hline
    \end{tabular}
    }
    \caption{Model Performance Comparison for GPT-3.5 and DSPY. We bold the highest final validation rating on equivalent models (GPT 3.5) between our method and DSPY. Our method outperforms DSPY on all use cases and by an average of 6\%.}
    \label{tab:gpt35_dspy_performance}
\end{table*}

\subsection{Gradient Summarization}
While methods like \citep{auto_prompt_optim_with_gradient_descent} use the gradients individually to then generate new prompts (ie. 5 gradients would lead to generating 5 new prompts), we find that this leads to prompts that are far too specific to certain questions.
Additionally, the task of evaluating these candidate prompts consumes API calls that can lead to costly compute bills.
Therefore, we found the most effective way of generating new general-purpose prompts was to summarize all gradient feedback into one general paragraph that could apply to the dataset as whole.
We feed the task description and the gradients computed in the previous step to the gradient summarization module and use an LLM to generate a one-paragraph summary of the critiques taking into consideration the task at hand.
This step can be thought of as analogous to averaging gradients over a mini-batch to stabilize training.

\subsection{Prompt Editor}
Our prompt editor module then takes in the current prompt, the summarized gradient, and the task description and outputs a candidate prompt for each prompt within our beam (doubling our beam size momentarily).
The new prompts should likely address the critiques provided in the previous iteration.
Each candidate prompt is checked to ensure every slot in the current prompt is present in the new prompt in order to avoid information loss of any kind.
We then perform an evaluation of each new candidate prompt.
Specifically, we take a randomly sampled subset (five rows) of the dataset and feed the responses to our generate and evaluate modules, choosing the two candidate prompts with the highest average rating.
As our beam is now 5 prompts wide we reduce to a beam size of 3 through selecting the top 3 performers as given by the upper confidence bound \citep{ucb}.
This reduction step thus always retains the best-performing prompt from previous iterations, reducing the variance from evaluating candidate prompts on a small sample size.

\begin{figure*}[ht]
    \centering
    \includegraphics[width=\textwidth]{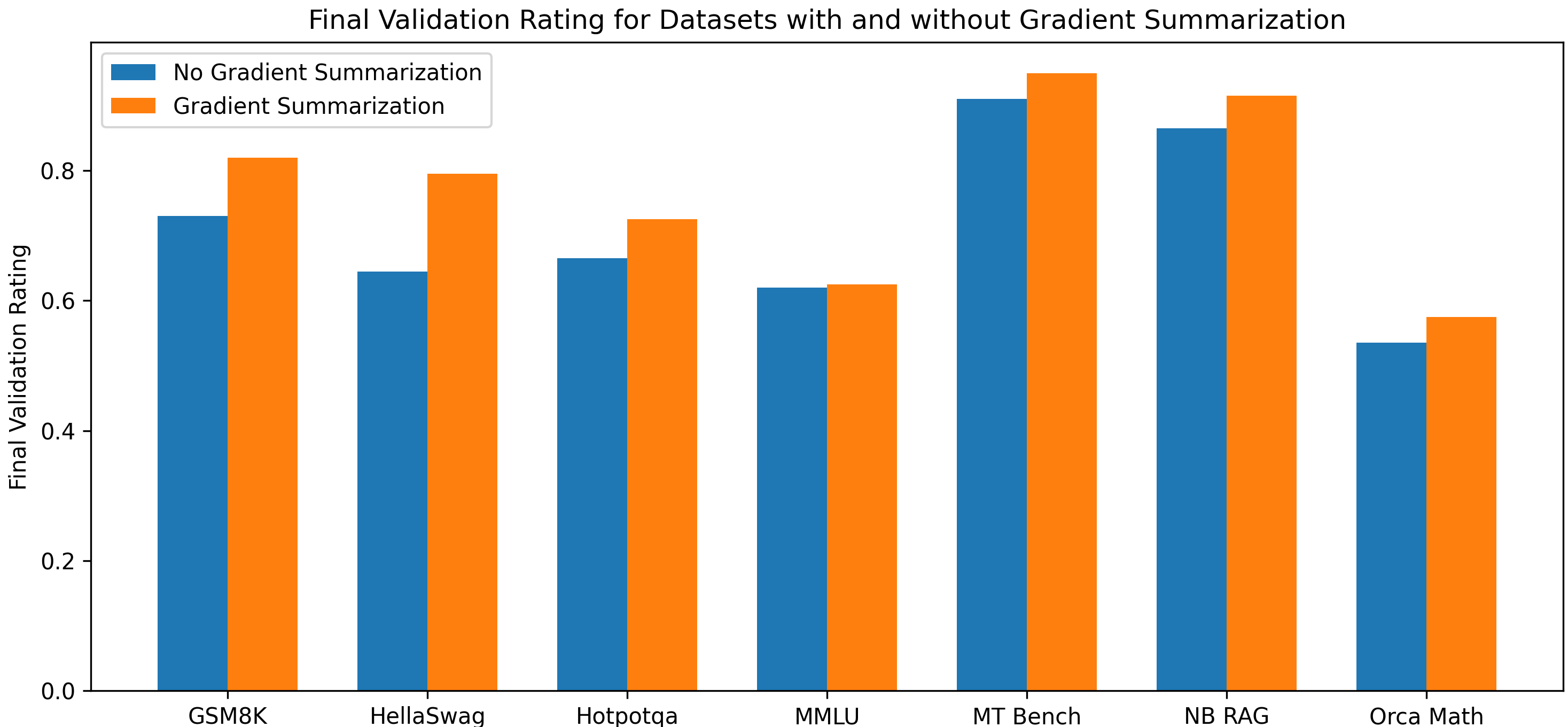}
    \caption{Our gradient summarization approach outperforms no gradient summarization by 5\%.}
    \label{fig:gradient_summ}
\end{figure*}

\section{Experiments \& Results}
In order to benchmark our method we utilize common datasets: GSM8k \citep{GSM8K}, Orca Math \citep{orca_math}, Neural Bridge RAG \citep{neural_bridge}, Hella Swag \citep{hellaswag}, HotPot QA \citep{hotpotqa}, MMLU \citep{mmlu} and a mixture of the first turn from MT Bench and Vicuna Bench \citep{vicuna_bench}.
We do not use any expected answers in our MT Bench and Vicuna Bench results, highlighting our use of user-provided evaluation criteria extends our technique to datasets without expected answers.
For each dataset if there is an available training dataset we extract 30 random samples for our training set and 200 random samples from the validation or test set in order to simulate typical industry evaluation flows.
If there is no available test or validation set we sample 200 from the train set and hold them out during training.
10 iterations are run over the train set with 10 rows being evaluated each iteration for each prompt (thus 30 calls for 3 candidate prompts) and the best prompt (the best performing prompt left in our beam), is extracted and used in our final validation.
All datasets used can found on our GitHub.\footnote{https://github.com/rungalileo/prompt\_optimization\_datasets}
Seeing as \citep{auto_prompt_optim_with_gradient_descent} is not directly comparable to our approach on these benchmarks due to their classification / exact-matching framework\footnote{Figure \ref{fig:gradient_summ} compares to the method introduced in \citep{auto_prompt_optim_with_gradient_descent} after extending their work to use LLM-as-a-judge metrics, showcasing the gradient summarization module improves upon their approach.}, we instead compare to the popular prompt optimizer, DSPY \citep{dspy}.
As DSPY is an abstracted prompting tool that does not allow users to directly control the initial prompt, we extracted DSPY's initial prompt and used it as the starting point for our method to ensure a fair comparison.
The same evaluation criteria is used with DSPY and we create a custom LLM-as-a-judge metric within their system in order to make our methods as comparable as possible.
We utilize the DSPY COPRO optimizer and allow for 10 passes over the dataset to ensure a fair comparison.
For our method we conduct evaluation, gradient generation and summarization with GPT 3.5 (`gpt-3.5-turbo-0125').
Prompt editing is performed with GPT4o (`gpt-4o-2024-05-13') in order to ensure high-quality new prompts.
The initial prompts, the final prompts, the task descriptions and evaluation criteria used for the respective tasks are provided in the appendix.

As you can see in Table \ref{tab:gpt35_dspy_performance} our method improves and outperforms DSPY on all use cases. 
We average an improvement of 14\% over our initial prompts, showcasing the robustness and strength of our optimizer.

\paragraph{Gradient Summarization}
We also conduct an ablation where we remove the summarization module in order to assess it’s efficacy.
Removing the summarization module is analogous to comparing our method with \citep{auto_prompt_optim_with_gradient_descent} after extending their method to use LLM-as-a-judge evaluation.
We note an average validation increase of 5\% with the introduction of the summarization module.
Upon examining the HotPot QA final prompt, it became obvious the prompt became far too specific to single gradients found throughout the training loop.
Below is an extract from the final HotPot QA prompt that clearly attaches to information relevant to individual examples:
\begin{quote}
    ``Reasoning: Let's think step by step to logically deduce the answer. Focus on the relevant information in the context to arrive at the correct answer. Ensure you consider all details, specific terms, and names mentioned in the context, such as the regiment within the Guards Division, the term 'shorts,' the founding year of Image Comics, middle names, the specific crime Brian Nichols was on trial for, and the full names of individuals. Provide a comprehensive response covering all individuals mentioned in the context where applicable."
\end{quote}
Thus, we find that the gradient summarization module is essential for maintaining high-quality prompts that can generalize to a validation dataset.

\section{Conclusion}
We introduce a scalable, flexible prompt optimization framework that improves LLM performance by an average of 14\% across many popular datasets.
In all scenarios our framework is also able to find a better prompt than our initial prompt.
Finally, our framework can be used across many of scenarios including those without expected answers, illustrating the flexibility of our framework.
By automating the prompt engineering process and demonstrating consistent performance gains across multiple datasets and models, our work enables efficient, flexible and scalable utilization of LLMs in real-world applications, reducing the time and effort required to achieve high-quality outputs.

\section{Limitations}
Our system currently supports only LLM-as-a-judge metrics. 
While these metrics are effective in many scenarios, they may not be suitable for all tasks. 
Expanding the system's capability to support any user-defined metric, including numerical and domain-specific metrics, is left for future work.
The use of LLMs for evaluation also introduces potential biases inherent in the models themselves. This has the potential to affect the accuracy of the evaluation process. 
Finally, there is still a dependence on the user to define the task descriptions and evaluation criteria, which have a large impact on final prompt quality. 
Streamlining these aspects to require minimal user input while maintaining effectiveness remains a challenge.

\newpage

\bibliography{custom}
\newpage
\newpage
\appendix
\section{Prompt Optimization Results}
For each dataset we list out the initial prompt, the task description, the evaluation criteria and the final prompt obtained. We provide the final prompt for all methods: GPT 3.5 with DSPY, and GPT 3.5 with our method.

\subsection{GSM8k}
\textbf{Task Description:}\newline The task is answering grade school math questions.
\newline
\newline
\textbf{Evaluation Criteria:}\newline Does the output align with the expected answer? \
The questions are math questions. \
Check if the answer matches the expected answer. Give it a 1 if a math teacher \
would consider the answer correct. Give it a 0 if the answer is incorrect.\
Do not worry about intermediate calculations, only the final answer.
\newline
\newline
\textbf{Initial prompt for our method \& DSPY:}\newline 
The task is to answer math questions.\newline\newline---\newline\newline Follow the following format.\newline\newline Question: 
\begin{verbatim}
${question}
\end{verbatim}
Reasoning: Let's think step by step in order to 
\begin{verbatim}
${produce the answer}
\end{verbatim}
We ...\newline
Answer: the answer to the question provided\newline\newline---\newline\newline Question: \{question\}\newline
Reasoning: Let's think step by step in order to Answer: 
\begin{verbatim}
${answer}
\end{verbatim}
Answer:
\newline
\newline
\textbf{Best Prompt GPT 3.5 (DSPY):}\newline
Follow the following format. \newline \newline Question: 
\begin{verbatim}
${question}
\end{verbatim}
Reasoning: Let's think step by step in order to 
\begin{verbatim}
${produce the answer}
\end{verbatim}
We ... \newline
Step-by-Step Explanation:» the answer to the question provided \newline \newline --- \newline \newline Question: \{question\} \newline
Reasoning: Let\'s think step by step in order to find the solution. We will break down the equation and solve for the unknown variable. Step-by-Step Explanation: [Insert detailed steps and calculations here] \newline
Step-by-Step Explanation:»
\newline
\newline
\textbf{Best Prompt GPT 3.5 (Our method):} 
\newline
The task is to answer grade school math questions accurately and clearly. Follow these detailed steps to ensure clarity and correctness in your response. \newline \newline Question: 
\begin{verbatim}
{question}
\end{verbatim}
Reasoning: Let's think through the problem step by step to produce the correct answer. Follow these instructions carefully: \newline \newline 1. Identify Key Information: \newline - Read the 
\begin{verbatim}
{question}
\end{verbatim}
carefully. \newline - Identify and list the key pieces of information given. \newline - Clearly state what the question is asking for. \newline \newline 2. Choose the Appropriate Method: \newline - Determine which mathematical methods or formulas are needed to solve the problem. \newline - Explain why this method or formula is appropriate for the given question. \newline \newline 3. Apply the Method: \newline - Carefully apply the chosen method or formula to the problem. \newline - Show each step of your calculation clearly and in order. \newline - Explain your reasoning at each stage to ensure clarity. \newline \newline 4. Double-Check: \newline - After arriving at a solution, double-check your calculations to confirm their accuracy. \newline - Review your explanation to ensure it is logical and easy to understand. \newline - Revisit the question to make sure your solution addresses exactly what was asked. \newline \newline Answer:

\subsection{Orca Math}
\textbf{Task Description}: \newline
The task involves solving math word problems.
\newline
\newline
\textbf{Evaluation Criteria}:
\newline
Does the output align with the expected answer? \
The questions are math questions. \
Check if the answer matches the expected answer. Give it a 1 if a math teacher \
would consider the answer correct. Give it a 0 if the answer is incorrect.\
Do not worry about intermediate calculations, only the final answer.
\newline
\newline
\textbf{Initial prompt for our method \& DSPY:}\newline 
The task is to answer math questions.\newline\newline---\newline\newline Follow the following format.\newline\newline Question: 
\begin{verbatim}
${question}
\end{verbatim}
Reasoning: Let's think step by step in order to 
\begin{verbatim}
${produce the answer}
\end{verbatim}
We ...\newline
Answer: the answer to the question provided\newline\newline---\newline\newline Question: \{question\}\newline
Reasoning: Let's think step by step in order to Answer: 
\begin{verbatim}
${answer}
\end{verbatim}
Answer:
\newline
\newline
\textbf{Best Prompt GPT 3.5 (DSPY): }\newline
Take advantage of the large size of the language model and generate detailed explanations for the math questions instead of just the answers.» \newline \newline --- \newline \newline Follow the following format. \newline \newline Question: 
\begin{verbatim}
${question}
\end{verbatim}
Reasoning: Let's think step by step in order to 
\begin{verbatim}
${produce the answer}
\end{verbatim}
We ... \newline
Explanation: the answer to the question provided \newline \newline --- \newline \newline Question: \{question\} \newline
Reasoning: Let's think step by step in order to
\newline
\newline
\textbf{Best Prompt GPT 3.5 (Our method)}: 
\newline
The task involves solving math word problems accurately by following a clear and logical reasoning process. Please adhere to the following format to ensure a coherent and cogent response: \newline \newline --- \newline \newline Question: 
\begin{verbatim}
{question}
\end{verbatim}
Reasoning: Let's think step by step in order to produce the answer. We will break down the problem into smaller parts, perform the necessary calculations, and show all work and thought processes clearly. Ensure each step logically follows from the previous one and aligns with the evaluation criteria of correctness, completeness, and clarity. \newline \newline Answer: Provide the final answer to the question based on the reasoning process above. \newline \newline --- \newline \newline Question: \{question\} \newline
Reasoning: Let's think step by step in order to produce the answer. We... \newline
Answer:

\subsection{Neural Bridge RAG}
\textbf{Task Description}: \newline
The task is a question answering task given specific context that should have the answer.
\newline
\newline
\textbf{Evaluation Criteria}: 
\newline 
Does the llm output match the expected answer? \
If the model says it does not have enough context to answer the question give it a 0. \
Otherwise judge whether a human would grade the output as matching the expected answer. \
Adding context around the answer is fine as long as the answer is correct according to the expected answer within the brackets <EXPECTED ANSWER>. \
If it does match give it a 1. If it does not give it a 0.
\newline
\newline
\textbf{Initial prompt for our method \& DSPY}:\newline
The task is a question answering task given context that should have the answer. \newline \newline --- \newline \newline Follow the following format. \newline \newline Question: 
\begin{verbatim}
${question}
\end{verbatim}
Context: 
\begin{verbatim}
${context}`
\end{verbatim}
Reasoning: Let's think step by step in order to 
\begin{verbatim}
${produce the answer}
\end{verbatim}
We ... \newline \newline Answer: the answer to the question given the context provided \newline \newline --- \newline \newline Question: \{question\} \newline
Context: \{context\} \newline
Reasoning: Let's think step by step in order to
\newline
\newline
\textbf{Best Prompt GPT 3.5 (DSPY)}: 
\newline
Guide the model to closely examine the question through context, utilize advanced inference techniques to deduce implicit information effectively, and generate a comprehensive and sophisticated response geared towards enhancing overall understanding and insights.» \newline \newline --- \newline \newline Follow the following format. \newline \newline Question: 
\begin{verbatim}
${question}
\end{verbatim}
Context: 
\begin{verbatim}
${context}
\end{verbatim}
Reasoning: Let's think step by step in order to 
\begin{verbatim}
${produce the answer}
\end{verbatim}
We ... \newline
Sophisticated Answer: the answer to the question given the context provided \newline \newline --- \newline \newline Question: \{question\} \newline
Context: \{context\} \newline
Reasoning: Let's think step by step in order to
\newline
\newline
\textbf{Best Prompt GPT 3.5 (Our method)}: 
\newline
You are tasked with answering a question based on a given context. Follow the detailed steps below to ensure your response is specific, comprehensive, and derived solely from the context provided. \newline \newline 1. Carefully read and fully understand the question. Clarify exactly what information is being sought by the question. \newline 2. Thoroughly analyze the given context. Identify and extract all relevant pieces of information that directly pertain to the question. \newline 3. Think logically and proceed step-by-step to connect the extracted details from the context to the question. Clearly explain your reasoning process, ensuring it is detailed, coherent, and directly related to the context. \newline 4. Provide the final answer based on your detailed reasoning process, ensuring it directly addresses the question. \newline \newline Use the following format for your response: \newline \newline Question: 
\begin{verbatim}
{question}
\end{verbatim}
Context: 
\begin{verbatim}
{context}
\end{verbatim}
Reasoning: Let's think step by step to understand and find the answer. First, identify specific details from the context that are relevant to the question. Then, logically connect these details to derive the answer. Ensure your reasoning is clear, detailed, and coherent. \newline \newline Answer: The answer to the question based on the context provided

\subsection{HellaSwag}
\textbf{Task Description}: \newline The task is to complete a sentence with the most logical of 4 possible options.
\newline
\newline
\textbf{Evaluation Criteria}: \newline Does the output match the the expected answer? \
If it does give it a 1. If it does not give it a 0.  \
It does not have to match exactly but it should be close enough that a reasonable human would consider the output to match the expected answer. \
Make sure that the chosen completion is the correct completion.
\newline
\newline
\textbf{Initial prompt for our method \& DSPY}: 
\newline
The task is to complete a sentence with the most logical of 4 possible options. \newline \newline --- \newline \newline Follow the following format. \newline \newline Context: 
\begin{verbatim}
Context for the question
\end{verbatim}
Ending 1: 
\begin{verbatim}
First possible ending for the question
\end{verbatim}
Ending 2: 
\begin{verbatim}
Second possible ending for the question
\end{verbatim}
Ending 3: 
\begin{verbatim}
Third possible ending for the question
\end{verbatim}
Ending 4: 
\begin{verbatim}
Fourth possible ending for the question
\end{verbatim}
Reasoning: Let's think step by step in order to 
\begin{verbatim}
${produce the answer}
\end{verbatim}
We ... \newline \newline Answer: the answer to the question given the possible endings provided \newline \newline --- \newline \newline Context: \{context\} \newline
Ending 1: \{ending1\} \newline
Ending 2: \{ending2\} \newline
Ending 3: \{ending3\} \newline
Ending 4: \{ending4\} \newline
Reasoning: Let's think step by step in order to
\newline
\newline
\textbf{Best Prompt GPT 3.5 (DSPY)}: The task is to complete a sentence with the most logical of 4 possible options. \newline \newline --- \newline \newline Follow the following format. \newline \newline Context: 
\begin{verbatim}
Context for the question
\end{verbatim}
Ending 1: 
\begin{verbatim}
First possible ending for the question
\end{verbatim}
Ending 2: 
\begin{verbatim}
Second possible ending for the question
\end{verbatim}
Ending 3: 
\begin{verbatim}
Third possible ending for the question
\end{verbatim}
Ending 4: 
\begin{verbatim}
Fourth possible ending for the question
\end{verbatim}
Reasoning: Let's think step by step in order to 
\begin{verbatim}
${produce the answer}
\end{verbatim}
We ... \newline \newline Answer: the answer to the question given the possible endings provided \newline \newline --- \newline \newline Context: \{context\} \newline
Ending 1: \{ending1\} \newline
Ending 2: \{ending2\} \newline
Ending 3: \{ending3\} \newline
Ending 4: \{ending4\} \newline
Reasoning: Let's think step by step in order to
\newline
\newline
\textbf{Best Prompt GPT 3.5 (Our method)}: \newline
Your task is to complete a sentence with the most logical of 4 possible options. Follow these detailed instructions to ensure the best possible output. \newline \newline\#\#\# Task Instructions \newline \newline 1. Understand the Context: Carefully read the provided context to grasp the situation, the main idea, and any underlying nuances. Pay close attention to the details that are crucial for understanding the sentence. \newline 2. Analyze Each Option: Evaluate each possible ending individually. Consider its relevance, logical consistency, and coherence with the context. Think about how each option fits with the main idea and details provided in the context. \newline 3. Compare Against the Context: Cross-check each ending with the context to ensure it fits seamlessly and supports the overall meaning. Eliminate any options that do not make sense or disrupt the flow of the sentence. \newline 4. Reasoning Process: Explain your thought process step-by-step. Consider the context and evaluate each ending thoroughly. Justify why each ending is or isn't suitable, providing clear reasons for your choices. \newline 5. Select the Most Logical Ending: Choose the ending that best completes the sentence in a coherent and meaningful way. Ensure that the chosen ending aligns perfectly with the context and enhances the overall understanding of the sentence. \newline \newline \#\#\# Format \newline \newline Context: 
\begin{verbatim}
{context}
\end{verbatim}
Ending 1: 
\begin{verbatim}
{ending1}
\end{verbatim}
Ending 2: 
\begin{verbatim}
{ending2}
\end{verbatim}
Ending 3: 
\begin{verbatim}
{ending3}
\end{verbatim}
Ending 4: 
\begin{verbatim}
{ending4}
\end{verbatim}
Reasoning: Let's think step-by-step to produce the answer. We need to consider the context and evaluate each ending one by one to determine which is the most logical. We should look for consistency, relevance, and coherence in the story. \newline \newline Answer: The answer to the question given the possible endings provided. \newline \newline --- \newline \newline Context: \{context\} \newline
Ending 1: \{ending1\} \newline
Ending 2: \{ending2\} \newline
Ending 3: \{ending3\} \newline
Ending 4: \{ending4\} \newline
Reasoning: Let's think step-by-step in order to...\newline

\subsection{Hotpot QA}
\textbf{Task Description}:\newline
The task is to reason over context given a question that will require logical deduction.
\newline
\newline
\textbf{Evaluation Criteria}: \newline
Does the llm output answer the question correctly while \
only the context provided? Ensure that the model adheres to the context while providing a correct answer. \
If it does give it a 1. If it does not give it a 0.  \
If the answer is that the context is not provided give the answer a 0. 
\newline
\newline
\textbf{Initial prompt for our method \& DSPY}:\newline
The task is a question answering task given context that should have the answer. \newline \newline --- \newline \newline Follow the following format. \newline \newline Question: 
\begin{verbatim}
${question}
\end{verbatim}
Context: 
\begin{verbatim}
${context}
\end{verbatim}
Reasoning: Let's think step by step in order to 
\begin{verbatim}
${produce the answer}
\end{verbatim}
We ... \newline \newline Answer: the answer to the question given the context provided \newline \newline --- \newline \newline Question: \{question\} \newline
Context: \{context\} \newline
Reasoning: Let's think step by step in order to
\newline
\newline
\textbf{Best Prompt GPT 3.5 (DSPY):} \newline
Follow the following format. \newline \newline Question: 
\begin{verbatim}
${question}
\end{verbatim}
Context: 
\begin{verbatim}
${context}
\end{verbatim}
Reasoning: Let's think step by step in order to 
\begin{verbatim}
${produce the answer}
\end{verbatim}
We ... \newline \newline ANSWER: the answer to the question given the context provided \newline \newline --- \newline \newline Question: \{question\} \newline
Context: \{context\} \newline
Reasoning: Let's think step by step in order to
\newline
\newline
\textbf{Best Prompt GPT 3.5 (Our method)}: 
\newline
You need to address a question using the given information through logical reasoning. \newline \newline --- \newline \newline Use the structure below to deliver a comprehensive and precise answer. \newline \newline Question: 
\begin{verbatim}
{question}
\end{verbatim}
Context: 
\begin{verbatim}
{context}
\end{verbatim}
Reasoning: Let's break this down systematically to arrive at the solution. Begin by pinpointing the essential points in the context relevant to the question. Clearly enumerate each critical point. Then, interpret these points logically, elaborating on how each one aids in establishing a link between the context and the question. Ensure careful consideration of how each point bolsters your argument. Conclusively, infer the answer based on this logical link, ensuring your explanation is consistent, detailed, and specifically addresses the question. \newline \newline Answer: the answer to the question given the context provided

\subsection{MMLU}
\textbf{Task Description:} \newline
The task is to choose the correct answer from a list of possible answers on a variety of knowledge questions.
\newline
\newline
\textbf{Evaluation Criteria:} \newline Does the output match the the expected answer? \
If it does give it a 1. If it does not give it a 0.  \
It does not have to match exactly but it should be close enough that a reasonable human would consider the output to match the expected answer. \
Make sure that the chosen completion is the correct completion.
\newline
\newline
\textbf{Initial Prompt for our method \& DSPY} \newline 
The task is a question answering task given specific context that should have the answer. \newline \newline --- \newline \newline Follow the following format. \newline \newline Question: 
\begin{verbatim}
${question}
\end{verbatim}
Choices: 
\begin{verbatim}
${choices}
\end{verbatim}
Reasoning: Let's think step by step in order to 
\begin{verbatim}
${produce the answer}
\end{verbatim}
We ... \newline \newline Answer: the answer to the question given the context provided \newline \newline --- \newline \newline Question: \{question\} \newline
Choices: \{choices\} \newline
Reasoning: Let's think step by step in order to
\newline
\newline
\textbf{Best Prompt GPT 3.5 (DSPY)}: \newline
Follow the following format. \newline \newline Question: 
\begin{verbatim}
${question}
\end{verbatim}
Choices: 
\begin{verbatim}
${choices}
\end{verbatim}
Reasoning: Let's think step by step in order to 
\begin{verbatim}
${produce the answer}
\end{verbatim}
We ... \newline \newline Acquire comprehensive knowledge by exploring the prompt» the answer to the question given the context provided \newline \newline --- \newline \newline Question: \{question\} \newline
Choices: \{choices\} \newline
Reasoning: Let's think step by step in order to determine the correct response. We need to carefully analyze the information provided in the question and consider all possible options before making a decision. By breaking down the question and examining each choice thoughtfully, we can arrive at the most accurate answer. This approach will not only help us select the correct response but also deepen our understanding of the topic at hand. \newline
Acquire comprehensive knowledge by exploring the prompt»
\newline
\newline
\textbf{Best Prompt GPT 3.5 (Our method)}: \newline
The task is to choose the correct answer from a list of possible answers on a variety of knowledge questions. Follow the guidelines below to ensure a logical and accurate response. \newline \newline --- \newline \newline Follow this format: \newline \newline Question: 
\begin{verbatim}
`question`
\end{verbatim}
Choices: 
\begin{verbatim}
`choices`
\end{verbatim}
Reasoning: Let's think step by step in order to produce the answer. First, carefully read the question to understand what it is asking. Next, analyze the context and identify key pieces of information that relate to the question. Then, evaluate each choice using logical reasoning. Eliminate the choices that do not match the context or are less likely to be correct. Finally, select the best answer based on the analysis. \newline \newline Answer: the answer to the question given the context provided \newline \newline --- \newline \newline Example: \newline \newline Question: What is the capital of France? \newline \newline Choices: A) Berlin B) Madrid C) Paris D) Rome \newline \newline Reasoning: Let's think step by step in order to produce the answer. The question asks for the capital of France. From the context of general knowledge, we know that Berlin is the capital of Germany, Madrid is the capital of Spain, Paris is the capital of France, and Rome is the capital of Italy. Therefore, the correct answer must be Paris. \newline \newline Answer: C) Paris \newline \newline --- \newline \newline Now, let's proceed with the task. \newline \newline Question: \{question\} \newline Choices: \{choices\} \newline Reasoning: Let's think step by step in order to

\subsection{MT \& Vicuna Bench}
\textbf{Task Description:} \newline The task is to be a chat bot assistant that provides helpful answers to questions.
\newline
\newline
\textbf{Evaluation Criteria:} \newline Act as an impartial judge and evaluate the quality of the response provided by an \
AI assistant to the user question displayed below. Your evaluation should consider how helpful, thoughtful, \
informative and thorough an answer is. Only give perfect answers a 1.
\newline
\newline
\textbf{Initial Prompt for our method \& DSPY:} \newline 
The task is to be a chat bot assistant that provides helpful answers to questions. \newline \newline --- \newline \newline Follow the following format. \newline \newline Question: 
\begin{verbatim}
${question}
\end{verbatim}
Reasoning: Let's think step by step in order to 
\begin{verbatim}
${produce the answer}
\end{verbatim}
We ... \newline \newline Answer: the chat bot answer to the question posed by the user \newline \newline --- \newline \newline Question: \{question\} \newline Reasoning: Let's think step by step in order to
\newline
\newline
\textbf{Best Prompt GPT 3.5 (DSPY):} \newline
Engage with users in a friendly and informative manner to address their inquiries accurately and efficiently. \newline \newline --- \newline \newline Follow the following format. \newline \newline Question: 
\begin{verbatim}
${question}
\end{verbatim}
Reasoning: Let's think step by step in order to 
\begin{verbatim}
${produce the answer}
\end{verbatim}
We ... \newline 
\begin{verbatim}
```ChatBot Assistant:```
\end{verbatim}
the chat bot answer to the question posed by the user \newline \newline --- \newline \newline Question: \{question\} \newline Reasoning: Let's think step by step in order to
\newline
\newline
\textbf{Best Prompt GPT 3.5 (Our method):} \newline 
You are a chat bot assistant that provides helpful answers to questions in a personalized, engaging, and conversational tone. Follow the format below to ensure your responses are detailed, structured, and informative. Provide specific examples and offer in-depth analysis to enhance the overall quality of your answers. \newline \newline --- \newline \newline Question: \{question\} \newline \newline Reasoning: Let's think step by step to produce the answer. First, we ... [Provide a detailed, structured explanation, including specific examples where relevant to illustrate your points. Offer in-depth analysis and insights to enhance the overall informative value of your answer.] \newline \newline Answer: [Provide the final, clear, and concise answer to the question posed by the user.] \newline \newline --- \newline \newline Question: \{question\} \newline Reasoning: Let's think step by step in order to ...

\end{document}